\pdfoutput=1

\documentclass[11pt]{article}

\usepackage[]{acl}

\usepackage{times}
\usepackage{latexsym}

\usepackage{inconsolata}
\usepackage{comment}

\usepackage[T1]{fontenc}

\usepackage[utf8]{inputenc}

\usepackage{microtype}
\everypar={\looseness=-1}
\usepackage{makecell}
\usepackage[]{todonotes}
\usepackage{amsmath}
\usepackage{amsthm}

\theoremstyle{definition}

\usepackage{bbm}

\usepackage{array}
\newcolumntype{C}{>{$}c<{$}}
\newcolumntype{L}{>{$}l<{$}}

\usepackage{wrapfig}

\usepackage{todonotes}
\makeatletter
\newcommand*\iftodonotes{\if@todonotes@disabled\expandafter\@secondoftwo\else\expandafter\@firstoftwo\fi}  %
\makeatother

\newcommand{\defeq}[0]{\mathrel{\stackrel{\textnormal{\tiny def}}{=}}}

\usepackage{cleveref}
\crefname{section}{\S}{\S\S}
\Crefname{section}{\S}{\S\S}
\crefname{table}{Tab.}{}
\crefname{figure}{Fig.}{Figs.}
\crefname{algorithm}{Algorithm}{}
\crefname{equation}{Eq.}{Eqs.}
\crefname{line}{Line}{}
\crefname{appendix}{App.}{}
\crefformat{section}{\S#2#1#3}
\crefname{thm}{Theorem}{}
\crefname{cor}{Corollary}{}
\crefname{prop}{Proposition}{}
\crefname{def}{Definition}{}

\usepackage{amssymb,graphicx}
\usepackage{tikz}
\usepackage{pgfplots}
\usepackage{float}
\restylefloat{table}
\usepackage{booktabs}

\DeclareMathOperator*{\argmax}{argmax}
\newcommand*{\topK}{\mathrm{atop}_K}
\newcommand*{\bw}{\mathbf{w}}

\newcommand{\pairb}[1]{\left( #1 \right)}
\newcommand{\bigO}[1]{\mathcal{O}(#1)}

\newcommand{\xra}[1]{\overset{#1}{\rightsquigarrow}}

\newcommand{\defn}[1]{\textbf{#1}}

\newcommand{\scorep}[1]{s_p(#1)}

\newcommand{\nt}[1]{\mathrm{#1}}
\newcommand{\Xsigma}{\nt{X}_{\sigma}}
\newcommand{\Xnotsigma}{\nt{X}_{\overline{\sigma}}}
\newcommand{\Sstart}{\mathrm{S}}
\newcommand{\trees}{\mathcal{T}}
\newcommand{\spans}{\mathtt{spans}}

\newcommand{\citeposs}[1]{\citeauthor{#1}'s (\citeyear{#1})}

\title{A Structured Span Selector}
\author{Tianyu Liu~\;~Yuchen Eleanor Jiang~\;~Ryan Cotterell~\;~Mrinmaya Sachan \\
\setlength{\fboxsep}{2.5pt}%
\setlength{\fboxrule}{2.5pt}%
\fcolorbox{white}{white}{
  $\{$\texttt{\href{mailto:tianyu.liu@inf.ethz.ch}{tianyu.liu}, } \texttt{\href{mailto:yucjiang@ethz.ch}{yucjiang}, }\texttt{\href{mailto:ryan.cotterell@inf.ethz.ch}{ryan.cotterell}, }\texttt{\href{mailto:mrinmaya.sachan@inf.ethz.ch}{ mrinmaya.sachan}}$\}$\texttt{@inf.ethz.ch}
} \\
    {%
\setlength{\fboxsep}{2.5pt}%
\setlength{\fboxrule}{2.5pt}%
\fcolorbox{white}{white}{
    \includegraphics[width=.15\linewidth]{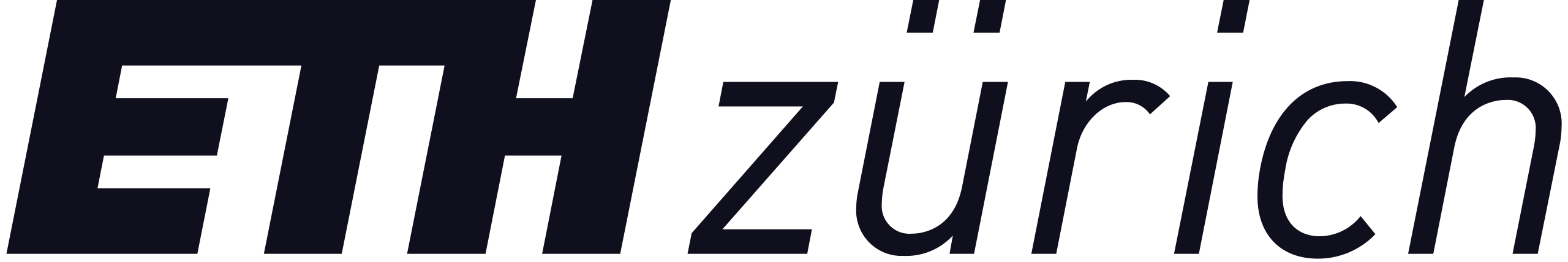}
}
}}

\begin{document}
\maketitle

\begin{abstract}
   Many natural language processing tasks, e.g., coreference resolution and semantic role labeling, require selecting text spans and making decisions about them. 
   A typical approach to such tasks is to score all possible spans and greedily select spans for task-specific downstream processing.
   This approach, however, does not incorporate any inductive bias about what sort of spans ought to be selected, e.g., that selected spans tend to be syntactic constituents.
   In this paper, we propose a novel grammar-based structured span selection model which learns to make use of the partial span-level annotation provided for such problems. 
   Compared to previous approaches, our approach gets rid of the heuristic greedy span selection scheme, allowing us to model the downstream task on an optimal set of spans. We evaluate our model on two popular span prediction tasks: coreference resolution and semantic role labeling.
   We show empirical improvements on both.
   \newline
\newline
\vspace{1.5em}
\hspace{.5em}\includegraphics[width=1.25em,height=1.25em]{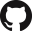}\hspace{.75em}\parbox{\dimexpr\linewidth-2\fboxsep-2\fboxrule}{\url{https://github.com/lyutyuh/structured-span-selector}}
\vspace{-.5em}
\end{abstract}

\section{Introduction}
The problem of selecting a continuous segment of the input text, termed a \defn{span}, is a common design pattern\footnote{A software engineering metaphor, which refers to a reusable solution to a commonly occurring problem.} %
in NLP.
In this work, we call this design pattern \defn{span selection}.
Common tasks that have a span prediction component include coreference resolution \cite{discourse_processing}, where the selected spans are mentions, semantic role labeling \cite{srl_palmer}, where the selected spans are arguments, question answering \cite{slp}, where the selected spans are answers, and named entity recognition \cite{ling_sp}, where the selected spans are entities.

\begin{figure} {
    \small
    \framebox[\width]{
    \begin{minipage}{0.98\linewidth}
   \paragraph{Coref:} \hspace{-9pt}
   [The President]\textsubscript{1} has said \big[[he]\textsubscript{1} and [his]\textsubscript{1} wife, now a [New York]\textsubscript{3} senator\big]\textsubscript{2} will spend weekends at \underline{
   [their]\textsubscript{2} house in \underline{
   Chappaqua}}
   .
   \\
    \paragraph{SRL:} \hspace{-4pt}
    [The most important thing about Disney]\textsubscript{1-\textsc{arg1}} \framebox{is\textsubscript{1}} [that [it]\textsubscript{2-\textsc{arg1}} \framebox{is\textsubscript{2}} [a global brand]\textsubscript{2-\textsc{arg2}} ]\textsubscript{1-\textsc{arg2}} .
    \end{minipage}
    }
}%
\caption{Examples of two span prediction tasks: \textit{Coreference} and \textit{SRL}. In Coref, $[s]_i$ denotes a span of text $s$ referring to the $i^{\text{th}}$ entity. In SRL, \framebox{i} denotes a set of predicates and $[s]_{i-*}$ denotes a set of arguments for the $i^{\text{th}}$ predicate.} \label{fig} \vspace{-2pt}
\end{figure}

In most of the tasks mentioned above, span selection is the first step.\footnote{Note that we also have joint models where we find optimal spans and make downstream decisions simultaneously \cite{lee-etal-2017-end,he-etal-2018-jointly}.}
After a set of candidate spans is determined, a classifier (often a neural network) is typically used to make predictions about the candidate spans. 
For instance, in coreference resolution, the selected spans (mentions) are clustered according to which entity they refer to; %
whereas in SRL, the spans (arguments) are classified into a set of roles. Two examples are shown in \cref{fig}. %

As the number of spans to consider in the input text can be quadratic in the length of the input,
candidate spans are greedily selected as potential antecedents, roles, or answers. %
While greedy span selection has become the de-facto approach in span prediction problems, it has several issues.
First, such approaches typically ignore the inherent structure of the problem. 
For example, spans of interest in problems such as coreference and SRL are typically syntactic constituents, an assumption supported by quantitative results.\footnote{For the OntoNotes dataset, \citet{he-etal-2018-jointly} report that 98.7\% of the arguments in SRL are constituents. For coreference, we find that 99.1\% of the mentions are constituents.} 
The lack of syntactic constraints on the spans of interest leads to a waste of computational resources, as all $\bigO{n^2}$ possible spans are enumerated by the model.

\begin{figure}
    \centering
    \includegraphics[width=\linewidth]{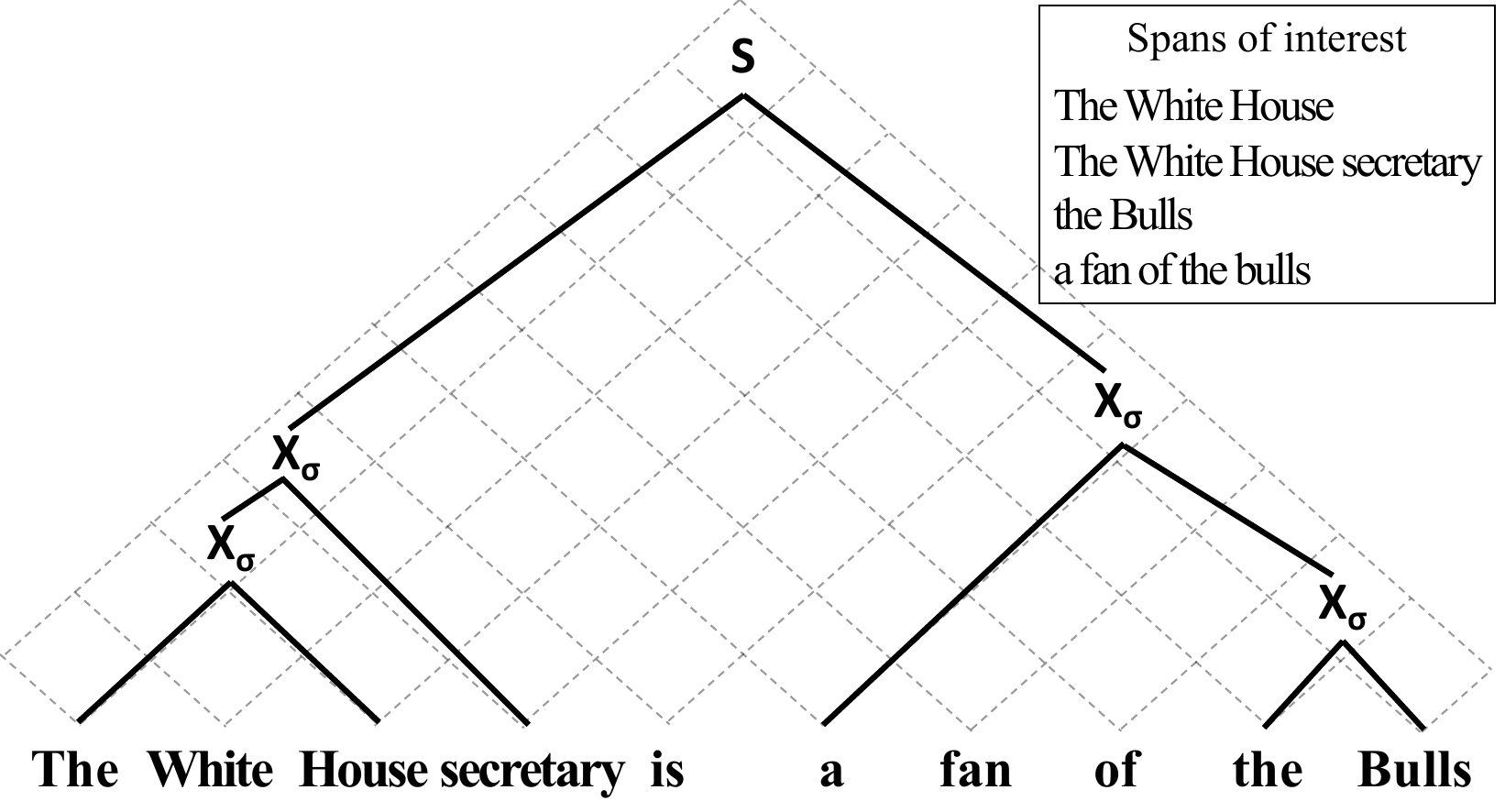}
    \caption{An example partial parse in our WCFG for coreference resolution and the set of spans it corresponds to. Production rules that do not involve the ``span of interest'' non-terminal ($\Xsigma$) are skipped as they do not affect the parsing result (see \cref{sec:grammar}). The traditional greedy approach considers $\bigO{n^2}$ spans denoted by the grid cells unless some pruning heuristics are applied, while the number of nonterminals in a CNF parse is $\bigO{n}$.\looseness=-1
        \setlength{\belowcaptionskip}{-10pt}
    }
    \label{fig:arch}
\end{figure}

In this paper, we propose a structured span selector for span selection.
Our span selector is a syntactically aware,  weighted context-free grammar that learns to score partial, possibly nested span annotations.
In the case of partial annotation, we marginalize
out the missing structure and maximize the marginal log-likelihood.
\cref{fig:arch} illustrates an example partial parse of our WCFG and the difference between the traditional greedy approach and our approach.\looseness=-1

We apply our span selector to both coreference and SRL.
In both cases, we optimize the log-likelihood of the joint distribution defined by the span selector and the conditional for the downstream task as defined in \newcite{lee-etal-2017-end} and \newcite{he-etal-2018-jointly}.
In contrast to previous approaches, which heavily rely on heuristics\footnote{\newcite{lee-etal-2017-end} consider spans such that the maximum span width is 10, the number of spans per word is 0.4, etc. Then, the spans are greedily pruned.} 
to prune the set of spans to be considered, our span selector  directly admits a tractable inference algorithm to determine the highest-scoring set of spans.
We observe that the number of spans our model selects is significantly lower than the number of spans considered in previous works, resulting in a reduction in the memory required to train these models.
Our approach leads to significant gains in both downstream tasks considered in this work: coreference and SRL. 
We find that our approach improves the performance of the end-to-end coreference model on the OntoNotes English dataset (0.4 F1) and the LitBank dataset (0.7 F1). 
On SRL, our model also achieves consistent gains over previous work. \looseness=-1

\section{Background and Related Work}
\subsection{Span Selection as a Design Pattern}
Many NLP tasks involve reasoning over textual spans, e.g., coreference resolution, semantic role labeling, named entity recognition, and question answering.
Models for these span prediction tasks often follow a common design pattern.
They decompose into two components: (i) a {\bf span selection} component where the model first selects a set of spans of interest, and (ii) a {\bf span prediction} component where a prediction (e.g., entity or role assignment) is made for the chosen set of spans.

As shown in previous papers \cite{zhang-etal-2018-neural-coreference,wu-etal-2020-corefqa}, the quality of the span selector can have a large impact on the overall model performance.
The span selector typically selects spans by selecting the start token and the end token in the span. Thus, there are inherently $n \choose 2$ textual spans, which is $\bigO{n^2}$, within a document of $n$ tokens to choose from. Previous span selection models \cite{lee-etal-2017-end, he-etal-2018-jointly} enumerate all possible spans in a brute-force manner and feed the greedily selected top-$k$ span candidates to the downstream prediction step. 

However, several span selection tasks require spans that are syntactic constituents, which is a useful, but often neglected \emph{inductive bias} in such tasks. 
For instance, in coreference resolution, mentions are typically noun phrases, pronouns, and sometimes, verbs. Similarly, in semantic role labeling, %
semantic arguments of a predicate are also typically syntactic constituents such as noun phrases, prepositional phrases, adverbs, etc. 
Our work uses a context-free grammar to enumerate spans.
The number of valid syntactic constituents in a constituency tree of a sequence of length $n$ is bounded by $\bigO{n}$, as the constituents can be viewed as the internal nodes of a binary parse tree in Chomsky normal form. Compared with brute-force enumeration and greedy pruning, this inductive bias provides us with a natural pruning strategy to reduce the number of candidate spans from $\bigO{n^2}$ to $\bigO{n}$ and provides us a more natural and linguistically informed way to model span-based tasks in NLP, through which we can employ parsing techniques and retrieve the optimal span selection for downstream tasks.

To further motivate our approach, we provide background on two popular span prediction tasks considered in our paper: coreference resolution and semantic role labeling (SRL). 
We also overview some previous papers on these tasks and contrast their methodology with ours.
Finally, we describe how our model can work with partial span-level annotations provided by datasets for these tasks.

\subsection{Coreference Resolution}
Most coreference resolution models involve two stages: mention detection and mention clustering. 
Traditional pipeline systems rely on parse trees and hand-engineered rules to extract mentions \cite{raghunathan-etal-2010-multi}. 
However, \citet{lee-etal-2017-end} show that we can directly detect mentions as well as assign antecedents to them in an end-to-end manner.\looseness=-1

In addition to this paper, other works have also explored better mention proposers for coreference. \citet{zhang-etal-2018-neural-coreference} use a multi-task loss to optimize the mention detector directly. \citet{swayamdipta-etal-2018-syntactic} also leverage syntactic span classification as an auxiliary task to assist coreference. \citet{thirukovalluru-etal-2021-scaling}, \citet{kirstain-etal-2021-coreference}, and \citet{dobrovolskii-2021-word} explore token-level representations to both reduce memory consumption and increase performance on longer documents. \citet{miculicich-henderson-2020-partially} and \citet{yu-etal-2020-neural} both improve the mention detector with better neural network structures. Yet they still need to manually set a threshold to control the number of selected candidate mentions and none of them could produce an optimal span selection for the downstream task. \citet{finkel-manning-2009-joint} situate NER in a parsing framework by explicitly incorporating named entity types into parse tree labels. In contrast, our work requires neither syntactic annotations nor hyperparameter tuning for mention selection.\looseness=-1

\subsection{Semantic Role labeling}

Semantic role labeling (SRL) extracts relations between predicates and their arguments. Two major lines of work in SRL are sequence-tagging models \cite{he-etal-2017-deep, marcheggiani-etal-2017-simple} and span-based models \cite{he-etal-2018-jointly,ouchi-etal-2018-span,li2019dependency}.
Sequence tagging models for SRL convert semantic role annotations to BIO sequences. The tagger generates a label sequence for one single predicate at a time. However, span-based models generate the set of all candidate arguments in one forward pass and classify their semantic roles with regard to each predicate. As discussed in \citet{he-etal-2018-jointly}, span-based models empirically perform better than sequence tagging models as they incorporate span-level features. Span-based models also do better at long-range dependencies as well as agreements with syntactic boundaries. Thus, we focus on span-based models in this work.

\subsection{Nested and Partial Span Annotations}\label{sec:nested}

Nested mentions and partially annotated mentions are two major concerns in this paper.
Most datasets for span prediction problems contain partial annotations of mentions. For example, singletons are not annotated in OntoNotes \cite{pradhan-etal-2012-conll}.
In the coreference resolution example given in \cref{fig}, the bracketed nested spans are annotated, while the underlined spans are valid mentions that are unannotated, %
since they do not co-refer with any other mention in the same document.
The same is also true in SRL. In the SRL example in \cref{fig}, there are two predicates (boxed words in the example) in one sentence. Their arguments are nested (i.e., \textsc{arg1} %
and \textsc{arg2} of the second predicate are located within \textsc{arg2} of the first predicate).\looseness=-1

\section{A Structured Span Selection Model}
In this section, we develop the primary contribution of our paper: A new model for span selection.
Specifically, we assert that almost all spans that a span selector should select are syntactic constituents; see \cref{fig} for two examples.
Under this hypothesis, a context-free grammar (CFG) 
is a natural model for span selection as spans selected by a CFG cannot overlap, i.e., every pair of spans selected by a CFG would either be nested or disjoint.

\subsection{Notation}
We first start by introducing some basic terminology.
We define a \textbf{document} $D$ as a sequence of sentences $\bw_1, \dots, \bw_{|D|}$.
Each \textbf{sentence} $\bw$ in the document is a sequence of words $[w_1, \dots, w_{|\bw|}]$.
A \textbf{span} is a contiguous subsequence of words in a sentence.
For instance, we denote the span from position $i$ to position $k$, i.e., $w_i \cdots w_k$, as $[i,k]$.

\subsection{Weighted Context-Free Grammars}\label{sec:grammar}

Next, we define weighted context-free grammars (WCFG), the formalism that we will use to build our span selector.
A WCFG is a five-tuple $\langle \Sigma, N, \Sstart, R, \rho \rangle$, where $\Sigma$ is an alphabet\footnote{An alphabet is a finite, non-empty set.} of terminal symbols, $N$ is a finite set of non-terminal symbols, $\Sstart \in N$ is the unique start symbol, $R$ is a set of production rules where a rule is of the form $\nt{X} \rightarrow \boldsymbol{\alpha}$ where $X \in N$ and $\boldsymbol{\alpha} \in \left(N \cup \Sigma\right)^*$, and $\rho : R \rightarrow \mathbb{R}_{\geq 0}$ is a scoring function that maps every production rule to a non-negative real number.\footnote{The scoring function can easily be generalized to map any production to a semiring value \citep{goodman-1999-semiring}.} %
We say a WCFG is in Chomksy normal form (CNF) if every production rule has one of three  forms: $\nt{X} \rightarrow \nt{Y}\, \nt{Z}$, where $\nt{X}, \nt{Y}, \nt{Z} \in N$, $\nt{X} \rightarrow x$, where $\nt{X} \in N$ and $x \in \Sigma$, or $\Sstart \rightarrow \varepsilon$.
For an input sentence $\bw$, a WCFG defines a weighted set of parse trees, which we will denote $\trees(\bw)$; we drop the argument from $\trees$ when the sentence is clear from the context.
\looseness=-1

We overload the scoring function $\rho$  to assign a weight to each parse tree $t \in \trees(\bw)$.
We define $\rho$ applied to a tree as follows
\begin{equation} \label{eq:parse_score}
    \rho\!\left(t\right) = \prod_{r \in t} \rho\!\left(r\right) \geq 0   
\end{equation} %
Given that $\rho$ returns a non-negative weight, our WCFG can be used to define a distribution over the set of all parses of a sentence 
\begin{equation}
    p(t \mid \bw) = \frac{\rho\!\left(t\right)}{Z}
\end{equation}
where $Z = \sum_{t \in \mathcal{T}(\bw)} \rho(t)$ is the sum of the scores of all parses. 
Using the familiar inside--outside %
algorithm \citep{baker1979trainable}, we can exactly compute $Z$ in $\bigO{|\bw|^3}$ time.\looseness=-1 %

\subsection{A WCFG Span Selector}
To convert from parse trees to spans, our paper exploits a simple fact: If our grammar is in CNF, %
then every parse tree $t \in \trees(\bw)$ corresponds to a \emph{unique} set of labeled spans.\footnote{The necessary and sufficient condition for this to hold is that the grammar has no unary or nullary rules, so every parse tree in the grammar can be written as a bracketed string. } %
Specifically, we write $[i, \nt{X}, k]$ if and only if the contiguous subsequence $w_i \cdots w_k$ corresponds to a constituent rooted at $\nt{X}$ in $t$.
We will denote the tree--span bijection as $\spans(\cdot)$ and write $M_t = \spans(t)$ to denote the set of spans $t$ implies.
We will also denote the set of all sets of spans viable under a CFG in CNF as $\mathcal{M}(\bw)$. \looseness=-1

To extract spans useful for downstream tasks,
we propose a simple WCFG.
The grammar has three non-terminals $N \defeq \{\Sstart, \Xsigma, \Xnotsigma \}$ where $\Sstart$ is the distinguished start symbol.
A span rooted at non-terminal $\Xsigma$, denoted $[i, \Xsigma, k]$, is termed a \defn{span of interest}; we will abbreviate $[i, \Xsigma, k]$ as $\sigma_{ik}$.
    Likewise, a span rooted at non-terminal $\Xnotsigma$, denoted $[i, \Xnotsigma, k]$, is termed a \defn{span of non-interest}; we will abbreviate $[i, \Xnotsigma, k]$ as $\overline{\sigma}_{ik}$. 
The full grammar is given in \cref{app:grammar}.
We define our  weight function $\rho: R \rightarrow \mathbb{R}_{\ge 0}$ as follows: %
\begin{equation}\label{eq:span-scorer}
    \rho\!\left({}_i\nt{X}_k \rightarrow {}_i \nt{Y}_j\,{}_j\nt{Z}_k\right) \defeq \begin{cases}
    \exp s_p\left(\sigma_{ik}\right) & \textbf{if } \nt{X} = \Xsigma \\
    1 & \textbf{otherwise}
    \end{cases}
\end{equation}
where $s_p$ is a learnable span-scorer that assigns a non-negative weight.
Note this definition of $\rho$ is an  \defn{anchored} scoring mechanism as it also makes use of the span indices $i$ and $k$.

Under the simplified scoring function in \cref{eq:span-scorer}, the score of a tree $t$ can be re-expressed as a product of the scores of the spans of interest in $\spans(t)$.
Specifically, for any tree $t$, we have
\begin{align}
    \rho(t) &= \prod_{r \in t} \rho(r) \label{eq:first-span}\\
            &= \exp \left( \sum_{\sigma_{ik}\in \spans(t)}  s_p\left(\sigma_{ik}\right)  \right) \label{eq:second-span} \\
            &\defeq \exp s_p\left(M_t\right)
\end{align}
where $M_t = \big\{ \sigma_{ik} \mid \sigma_{ik} \in \spans(t)\big\}$.
Note that the step from \cref{eq:first-span} to \cref{eq:second-span} follows from
the fact that only those spans rooted at $\Xsigma$ have a weight other than 1 under our choice of $\rho$ and, furthermore, $\rho$ ignores the body of the context-free rule. %
The problem of mention detection is hence converted from subset selection to finding an optimal parse tree that maximizes the score:
\begin{align} \label{eq:max_score_1}
    M^* &= \argmax_{M_t \in \mathcal{M}(\bw)} s_p(M_t) \\
    &= \spans\left( \argmax_{t \in \trees(\bw)} \rho(t) \right)  \label{eq:max_score_spanset}
\end{align}
where the ``Viterbi version'' of
the CKY algorithm (see \cref{app:cky}), yields an \textit{exact} algorithm for the $\argmax$ function in \cref{eq:max_score_spanset} that runs in $\bigO{|\bw|^3}$ time.\looseness=-1

Finally, in order to train our WCFG with only partial supervision, i.e., in the case when we do not observe the entire tree, we require the marginal probabilities of the spans of interest.
First, let $\mathcal{T}_{ik}$ be the set of parses that contain $\sigma_{ik}$.
Then, the marginal probability of the span of interest $\sigma_{ik}$ can be expressed as: \looseness=-1
\begin{align}
    p(\sigma_{ik} \mid \bw) = \sum_{t \in \mathcal{T}_{ik}} p(t \mid \bw)  
    = \sum_{t \in \mathcal{T}_{ik}} \frac{\rho(t)}{Z} 
\end{align}
As described by \citet{eisner-2016-inside}, we can compute $p(\sigma_{ik} \mid \bw)$ by computing
the derivative of the log-normalizer $\log Z$ with respect to $s_p(\sigma_{ik})$, i.e.,\looseness=-1
\begin{equation} \label{eq:anchored_prob}
    p(\sigma_{ik} \mid \bw) = \frac{\partial \log Z}{\partial s_p(\sigma_{ik})}
\end{equation}
A detailed derivation of \cref{eq:anchored_prob} can be found in \cref{app:ioisfb}. Automatic differentiation ensures that this marginal computation will have the \emph{same runtime} as the computation of $\log Z$ itself---to wit in $\bigO{|\bw|^3}$ time \citep{Griewank}. %

\section{Adaptations to Downstream Tasks}\label{sec:adaptation}

In this section, we introduce how our structured span selector can be applied in an end-to-end manner to coreference resolution and SRL. 

\subsection{Coreference Resolution}\label{sec:joint_coref_model}

The goal of coreference resolution is to link a span of interest $\sigma_{ij}$, termed a mention in the context of coreference resolution, to its \textbf{antecedent}.
Note that the antecedent is either another mention in the same document or the dummy antecedent,\footnote{Following \citep{lee-etal-2017-end} the dummy antecedent $\epsilon$ represents two possible
scenarios: (1) the span is not an entity mention or
(2) the span is an entity mention but it is not coreferent with any previous span.} which we denote as $\epsilon$.
We write $\sigma_{ij} \rightsquigarrow \sigma_{kl}$ to denote that $\sigma_{kl}$ is $\sigma_{ij}$'s antecedent.
When formulating coreference in a probabilistic manner, we have the following natural decomposition:
\begin{align}
     p(&\sigma_{ij}, \sigma_{ij} \rightsquigarrow \sigma_{kl}) \\
     &= \underbrace{p(\sigma_{ij})}_{\text{pr. } \sigma_{ij} \text{is a mention}} \times \quad \underbrace{p\left(\sigma_{ij} \rightsquigarrow \sigma_{kl} \mid \sigma_{ij} \right)}_{\text{pr. } \sigma_{ij}\text{'s antecedent is } \sigma_{kl}} \nonumber
\end{align}
The support of the above distribution is $\mathcal{X} \times \mathcal{Y}_{ij}$, where $\mathcal{X}$ is the set of all possible textual spans in $D$, $\mathcal{Y}_{ij} = \{\sigma_{kl} \mid j < l\} \cup \{\epsilon\}$ is the set of mentions preceding $\sigma_{ij}$ plus the dummy antecedent $\epsilon$ for every $\sigma_{ij} \in \mathcal{X}$.
In words, the above decomposition means that the probability of $\sigma_{ij}$ co-referring with the span $\sigma_{kl}$ is the 
probability of first recognizing that $\sigma_{ij}$ 
is itself a mention and then determining the link. %
In practice, this decomposition means that modelers can select $p(\sigma_{ij})$ and $p \left( \sigma_{ij} \rightsquigarrow \sigma_{kl}  \mid \sigma_{ij}\right)$ according to their taste and, importantly, independently of each other.
In this work, we explore treating $p(\sigma_{ij})$ as the WCFG span selector described in \cref{sec:grammar} and
$p\left(\sigma_{ij} \rightsquigarrow \sigma_{kl} \mid \sigma_{ij}\right)$ as \citeposs{lee-etal-2017-end} popular span ranking model for coreference.

\paragraph{\newcite{lee-etal-2017-end} as a Mention-Linker.}\label{sec:coref_e2e}

We now describe the mention-linker in \citet{lee-etal-2017-end}.
We define the mention-linker distribution as
\begin{align}
    p(\sigma_{ij} &\rightsquigarrow m \mid \sigma_{ij})  \\
    & =
    \frac{\exp s(\sigma_{ij}, m)}{\sum_{m' \in \mathcal{Y}_{ij}} \exp {s\pairb{\sigma_{ij}, m'}}} \nonumber
\end{align}
The scoring function $s(\cdot, \cdot)$ is defined in two cases: One case for $m = \epsilon$, the dummy antecedent, and one for $m = \sigma_{kl}$, a preceding span:
\begin{align}
    s(\sigma_{ij}, \epsilon) &= 0  \\
    s(\sigma_{ij}, \sigma_{kl}) &= s_m(\sigma_{ij}) + s_m(\sigma_{kl}) + s_a(\sigma_{ij}, \sigma_{kl}) \nonumber
\end{align}
The first score function, $s_m(\sigma_{ij})$, is a score for span $[i,j]$ being a mention.
The second function, $s_a \pairb{\sigma_{ij}, \sigma_{kl}}$, a score that $\sigma_{kl}$ is an antecedent of $[i,j]$. 
In this work, both $s_a$ and $s_m$ are computed by neural networks that take span representations as inputs.
However, in principle, they could be computed by any model.

\paragraph{Training.}
\citeposs{lee-etal-2017-end} model adopts a na\"ive greedy algorithm by taking the top $\lambda |D|$ spans with the highest mention scores $s_m$, where $\lambda$ is a hyperparameter that has to be manually defined for different datasets. 
However, finding a proper ratio $\lambda$ can be very tricky.
In contrast, in our setting we can optimize the final objective function which is the log-likelihood
of the joint distribution defined at the beginning of \cref{sec:joint_coref_model}:
\begin{equation}
    L_1 = \sum_{\bw \in D} \sum_{\sigma_{ij} \in \mathcal{G}_{\bw}} \!\!\log \! \sum_{m \in {\mathcal{G}}_{ij}} p(\sigma_{ij}, \sigma_{ij} \rightsquigarrow m) 
\end{equation}
where $\mathcal{G}$ is the (partially) annotated set of mentions, $\mathcal{G}_{\bw}$ is the set of all textual spans of $\bw$ in $\mathcal{G}$, and $\mathcal{G}_{ij}$ is the ground truth cluster that $\sigma_{ij}$ belongs to. %

\paragraph{Handling partial annotation with no singletons.}
Since in many coreference datasets, e.g., OntoNotes, only the mentions that are referred to more than once are annotated, %
learning a mention detector from such data requires the ability to handle the lack of singleton annotations.
To handle partial span annotations, we marginalize out
the unannotated singletons. 
This results in the following marginal log-likelihood
\begin{align}
    L_2 = \sum_{\bw \in D} \sum_{\sigma_{ij} \in \overline{\mathcal{G}}_{\bw}}  \log \Big( & p(\sigma_{ij} \rightsquigarrow \epsilon \mid \sigma_{ij}) p(\sigma_{ij})  \nonumber \\
    &  + (1-p(\sigma_{ij})) \Big) 
\end{align}
We optimize the loss $L = L_1 + L_2$ jointly.
Here, $\overline{\mathcal{G}}_{\bw}$ denotes the set of all spans of $\bw$ \emph{not} in $\mathcal{G}$.

\paragraph{Time Complexity.} %
For each sentence, the inside--outside algorithm \cref{eq:io} and CKY algorithm \cref{eq:cky} reduced to $\bigO{|\bw|}$ semiring matrix multiplications. 
Sentence-level parallelism can also be applied to all the sentences. 

\subsection{Semantic Role Labeling}\label{sec:srl_model}

The goal of SRL is to classify the \textbf{semantic role} of every argument $\sigma_{ij}$ with respect to a given predicate. 
Following the notation style in \cref{sec:joint_coref_model}, we use $\sigma_{ij} \xra{\ell} v$ to denote that $\sigma_{ij}$ has the semantic role $\ell$ in the frame of predicate $v$. The joint probability $p\left( \sigma_{ij}, \sigma_{ij} \xra{\ell} v \right)$ can be written as: %
\begin{align*}
     \underbrace{p(\sigma_{ij})}_{\text{pr. } \sigma_{ij} \text{ is an argument}} \times \quad \underbrace{p\left(\sigma_{ij} \xra{\ell} v \mid \sigma_{ij} \right)}_{\text{pr. } \sigma_{ij}\text{'s role is }  \ell \text{ w.r.t. predicate } v} %
\end{align*}
\paragraph{\newcite{he-etal-2018-jointly} as a Role Classifier.}\label{sec:srl_classifier}
As this work focuses on span selection, we choose the role classifier to be the popular and effective one from \citet{he-etal-2018-jointly}, and use gold predicates $v$ during training and evaluation.
The semantic role label $\ell$ takes its values from a discrete label space $\mathcal{L}$, which contains all the semantic roles plus the null relation $\epsilon$. The classifier then models the following probability distribution: 
\begin{align}
    & p\left(\sigma_{ij} \xra{\ell} v \mid \sigma_{ij} \right) = \frac{\exp {s(\sigma_{ij}, v, \ell)}}{\sum_{\ell' \in \mathcal{L}} \exp {s(\sigma_{ij}, v, \ell')}} \nonumber \\
    & s(\sigma_{ij}, v, \ell) = s_m(\sigma_{ij}) + s_{r}(\sigma_{ij}, v, \ell)
\end{align}
Similar to the coreference model of \newcite{lee-etal-2017-end}, $s_m(\sigma_{ij})$ is the score for span $[i,j]$ to be an argument and $s_{r}\pairb{\sigma_{ij}, v, \ell}$ for $[i,j]$ to play the role $\ell$ for predicate $v$.
Note that the score for the $\epsilon$ label (i.e., no relation), $s(\sigma_{ij}, v, \epsilon)$, is set to constant $0$, similar to the dummy antecedent case in coreference.\looseness=-1

\paragraph{Training Objective.}\label{sec:srl_training}
\citeauthor{he-etal-2018-jointly}'s model also suffers from the challenge of tuning the hyperparameter $\lambda$.
The training objective for SRL with our structured span selection model is then: 
\begin{equation} 
L_1 = \sum_{\sigma_{ij} \in \mathcal{G}_{\bw}} \sum_{v} \log{p(\sigma_{ij}, \sigma_{ij} \xra{\ell} v)} 
\end{equation}
where $\ell$ is the correct semantic label of $\sigma_{ij}$ with regard to predicate $v$.
Similar to coreference resolution, we handle the issue of partial annotation for the span selection model by adding the log-likelihood that $\sigma_{ij}$ may not be an annotated argument:\looseness=-1
\begin{align}
    L_2 = \sum_{\sigma_{ij} \in \overline{\mathcal{G}}_{\bw}}\sum_v  \log \Big( & p(\sigma_{ij} \xra{\epsilon} v \mid \sigma_{ij}) p(\sigma_{ij})  \nonumber \\ 
    &  + (1-p(\sigma_{ij})) \Big)
\end{align}
And the final objective function is $L = L_1 + L_2$.

\section{Experiments}
\subsection{The Greedy Baseline}
Previous work has mostly considered a greedy procedure for span selection as opposed to \cref{eq:max_score_spanset}. 
The approach produces a score $s_g(\sigma_{ik})$ \textit{independently} for each span $\sigma_{ik}$. 
As the number of spans $\sigma_{ik}$ is potentially very large, the set of spans is greedily pruned to a set of size $K$. 
For instance, in SRL, spans are selected for each sentence:%
\begin{align}
    & M^*_{\mathrm{topk}} \\
    &= \topK\Big( \Big\{ s_g(\sigma_{ik}) \mid 1 \leq i < k \leq |\bw| \Big\} \Big) \nonumber
\end{align}
However, in coreference resolution, a set of spans is selected for the entire document:
\begin{align}
     & M^*_{\mathrm{topk}}  \\
    &=\topK\Big( \bigcup_{\bw \in D} \hspace{-3pt} \Big\{ s_g(\sigma_{ik}) \mid 1 \leq i < k \leq |\bw| \Big\} \Big) \nonumber
\end{align}
where $\topK$ is shorthand for $\mathrm{argtop}_K$.
We will see in our experiments that tuning $K$ can be quite challenging (see \cref{fig:mention_detection}). %
Moreover, as the greedy approach scores each span independently, it ignores the structure of the provided span annotation. %

\subsection{Datasets} \label{sec:datasets}
\paragraph{Coreference.} 
We experiment on the CoNLL-2012 English shared task dataset (OntoNotes) \cite{pradhan-etal-2012-conll} and LitBank \cite{bamman-etal-2020-annotated} in our experiments.
As a part of our experiments on OntoNotes, we apply the speaker encoding in \citet{wu-etal-2020-corefqa}, that is using special tokens (\texttt{<speaker>}, \texttt{</speaker>}) to denote the speaker's name, as opposed to the original binary features used by \citet{lee-etal-2017-end}. 
This simple change brings a consistent boost to the performance by $0.2$ F1. 
A major difference between these two datasets is that LitBank has singleton mention annotations while OntoNotes does not. 
For LitBank, we use the standard 10-fold cross-validation setup, as is the standard practice.%

\paragraph{SRL.} We use the CoNLL-2012 SRL dataset. Gold predicates are provided to the model.

\subsection{Coreference Resolution}
\begin{table*}[ht!]
\centering
\resizebox{\textwidth}{!}{
\small
\begin{tabular}{lcccccccccc}
\toprule
                     & \multicolumn{3}{c}{MUC}     & \multicolumn{3}{c}{B$^3$}       & \multicolumn{3}{c}{CEAF$_{\phi_4}$}    &        \\ \cline{2-11} 
                     & P    & R    & F             & P    & R    & F             & P    & R    & F             &  Avg. F1      \\ \midrule
\citet{lee-etal-2017-end} & 78.4 & 73.4 & 75.8          & 68.6 & 61.8 & 65.0          & 62.7 & 59.0 & 60.8          & 67.2          \\
\citet{lee-etal-2018-higher} & 81.4 & 79.5 & 80.4          & 72.2 & 69.5 & 70.8          & 68.2 & 67.1 & 67.6          & 73.0          \\
\citeauthor{fei-etal-2019-end} & 85.4 & 77.9 & 81.4          & 77.9 & 66.4 & 71.7          & 70.6 & 66.3 & 68.4          & 73.8          \\
\citeauthor{kantor-globerson-2019-coreference} & 82.6 & 84.1 & 83.4          & 73.3 & 76.2 & 74.7          & 72.4 & 71.1 & 71.8          & 76.6          \\
\citet{joshi-etal-2019-bert} & 84.7 & 82.4 & 83.5          & 76.5 & 74.0 & 75.3          & 74.1 & 69.8 & 71.9          & 76.9          \\
\citet{joshi-etal-2020-spanbert} & 85.8 & 84.8 & 85.3          & 78.3 & 77.9 & 78.1          & 76.4 & 74.2 & 75.3          & 79.6          \\
\citeauthor{xu-choi-2020-revealing} & 85.7 & 85.3 & 85.5          & 78.6 & 78.6 & 78.6          & 76.8 & 74.8 & 75.8          & 79.9          \\\midrule
\citeauthor{joshi-etal-2020-spanbert} (S)                  & 86.6 & 84.5 & 85.6          & 80.4 & 77.3 & 78.8          & 77.8 & 74.0 & 75.8          & 80.1          \\
\texttt{Ours}                 & 86.1 & 85.5 & \textbf{85.8} & 79.8 & 78.8 & \textbf{79.3} & 77.4 & 75.4 & \textbf{76.4} & \textbf{80.5} \\ \bottomrule
\end{tabular}} 
\caption{Results on the CoNLL-2012 English shared task test set. Avg. F1 in the last column denotes the average F1 of MUC, B{$^3$}, and CEAF{$_{\phi_4}$}. \citeauthor{joshi-etal-2020-spanbert} (S) refers to the original end-to-end model with SpanBERT and trained with speaker encoding. The improvements shown in the table are significant under a two-tailed paired $t$-test.}
\label{tab:result_test_conll}
\end{table*}

We report the average precision, recall, and F1 scores of the standard MUC, B{$^3$}, CEAF{$_{\phi_4}$}, and the average CoNLL F1 score on the OntoNotes test set in \cref{tab:result_test_conll}. The average F1 scores on LitBank are shown in \cref{tab:result_test_litbank}.
For OntoNotes, we run the experiments with 5 random initializations and the improvements reported are significant under the two-tailed paired $t$-test. 

We compare our models with several representative previous works. 
In order to focus on comparing the impact of mention detection, we do not consider higher-order inference techniques in our models and report the non-higher order result from \citet{xu-choi-2020-revealing}. \citeauthor{joshi-etal-2020-spanbert} (S) is the major baseline that uses SpanBERT \cite{joshi-etal-2020-spanbert} and is trained with the speaker encoding discussed in \cref{sec:datasets}. 
This encoding yields an F1 score improvement of $0.2$ over the result reported by \citet{xu-choi-2020-revealing}.\looseness=-1

Our model with the structured mention detector achieves an F1 score of $80.5$, an improvement of $0.4$ F1 over the baseline. While on LitBank, our model achieves an F1 score of $76.3$, which is an improvement of a $0.7$ F1 over \citep{joshi-etal-2020-spanbert}. It can also be observed that this gain mainly comes from improved recall, which is because we have a superior mention detector that can retrieve mentions with better accuracy. We further analyze this result in the following section.\looseness=-1

\begin{table}[ht!]

\begin{center}
\begin{tabular}{cccc}
\toprule
          & Avg. P & Avg. R & Avg. F1       \\ \midrule
\citeauthor{bamman-etal-2020-annotated}    & -      & -      & 68.1          \\
\texttt{LB-MEM} & -      & -      & 75.7          \\
\texttt{U-MEM} & -      & -      & 75.9          \\ \midrule
\citeauthor{joshi-etal-2020-spanbert}    & 78.7  & 72.9   & 75.6          \\
\texttt{Ours}      & 77.4   & 75.3   & \textbf{76.3} \\ \bottomrule
\end{tabular}
\caption{Results on the test set of LitBank. The results are averaged over 10 train/dev/test splits. \texttt{LB-MEM} and \texttt{U-MEM} are reported in \citet{toshniwal-etal-2020-learning}.}
\label{tab:result_test_litbank}
\end{center}
\end{table}

\subsubsection{Analysis of Mention Detector}
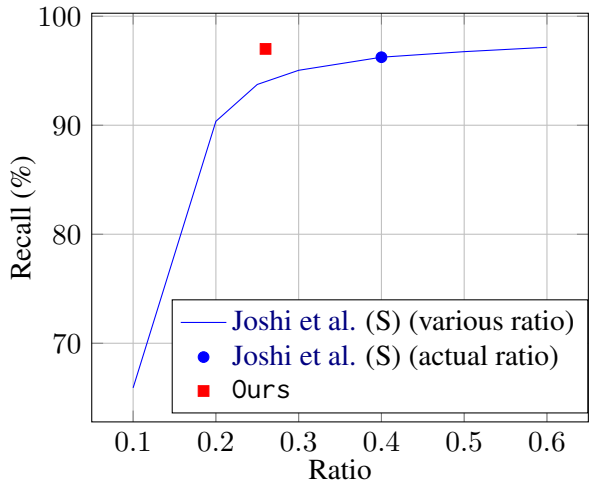
\begin{figure}
\begin{center}
\begin{tikzpicture}[scale=1]
    \begin{axis}[
        xlabel= Ratio,
        ylabel= Recall (\%),
        y label style={at={(axis description cs:0.05,0.5)}},
        x label style={at={(axis description cs:0.5,0.03)}},
        xmajorgrids=true,
        ymajorgrids=true,
        legend cell align={left},
        legend style={at={(1,0.3)}},
        height=7cm
    ]
      \addplot [mark=none,color=blue] 
      coordinates {
        (0.1,  65.91)
        (0.2,  90.36)
        (0.25, 93.74)
        (0.3,  95.04)
        (0.4, 96.24)
        (0.5, 96.75)
        (0.6, 97.15)
    };
    \addplot [mark=*,color=blue,only marks] 
      coordinates {
        (0.4,  96.24)
    };
    \addplot [mark=square*,color=red,only marks] 
      coordinates {
        (0.26,  97.0)
    };
    \legend{\citeauthor{joshi-etal-2020-spanbert} (S) (various ratio)\\\citeauthor{joshi-etal-2020-spanbert} (S) (actual ratio)\\ \texttt{Ours}\\}
    \end{axis}
\end{tikzpicture} 
\end{center}
\caption{Recall of gold mentions as we vary the ratio of spans kept. Ratio refers to the number of predicted mentions divided by $|D|$. Our mention detector significantly outperforms \citeauthor{joshi-etal-2020-spanbert} (S)'s with a ratio of $0.26$ and a recall of $97.0\%$. Yet \citeauthor{joshi-etal-2020-spanbert} (S) only achieves $96.2\%$ recall with a ratio of $0.40$.}
\label{fig:mention_detection}
\end{figure}

Next, we examine the performance of our proposed mention detection scheme. 
As shown in \cref{fig:mention_detection}, compared with \citeauthor{joshi-etal-2020-spanbert} (S), our model predicts mentions more accurately with a higher recall. 
In contrast to \citeauthor{joshi-etal-2020-spanbert} (S) who select $0.4|D|$ mention spans, our method on average selects $0.26|D|$.
The smaller span set makes the coreference model more efficient as well.

\subsubsection{Analysis of Structured Modeling}
To see how our structured modeling benefits coreference, we further compare our approach with a baseline \texttt{Sigmoid} which replaces the $p(\sigma_{ij})$ in \cref{eq:anchored_prob} with a simple non-structured estimator:
\begin{align}
    p_{\text{sigm}}(\sigma_{ij}) = \text{sigmoid} (s_p(\sigma_{ij})) 
\end{align}
where $\text{sigmoid}(x) = \frac{1}{1 + \exp(-x)}$. The loss function used for \texttt{Sigmoid} is the same as the structured model given in \cref{sec:joint_coref_model}. Through this comparison, we aim to show the effectiveness of structured modeling.
We also build a multi-task learning baseline \texttt{MTL} similar to \citet{swayamdipta-etal-2018-syntactic}. 
The baseline adds an auxiliary classifier that classifies spans into noun phrases, other syntactic constituents, or non-constituents. 
The coefficient of the multi-task loss is set to $0.1$ as in \citet{swayamdipta-etal-2018-syntactic}.\looseness=-1

This comparison is shown in \cref{tab:result_test_conll_sigmoid}. We find that replacing $p(\sigma_{ij})$ with unstructured $p_\text{sigm}(\sigma_{ij})$ 
degrades the performance by an F1 score of $0.3$. Thus, we conclude that the structured probability function is more expressive than $p_\text{sigm}(\sigma_{ij})$ as it  models the global annotation for each sentence. In contrast, the mention detectors in \citeauthor{joshi-etal-2020-spanbert} (S) and the \texttt{Sigmoid} model each span independently.

\subsubsection{Analysing the source of improvement}
Next, we try to explore where the gains of our model come from.

\paragraph{\textbf{Nested Mentions.}} We first investigate the capability of our structured model in handling nested mentions. \cref{tab:qualitative_depth} shows the recall rate of mentions of different nested depths. Here, nested depth refers to the level of nesting in the mentions. E.g., in the first example given in \cref{fig}, \textit{The president} is of depth 1, while \textit{he and his wife, now a New York senator} is of depth 2. 
As shown in \cref{tab:qualitative_depth}, the gains of our method are larger for deeply nested mentions, which highlight the capability of our structured span detector to handle more difficult nested mentions that cannot be handled by the greedy selector.

\begin{table}
\begin{center}
\begin{tabular}{cccc}
\toprule
Nested Depth    &  1    &  2        &  3+      \\ \midrule
\texttt{Greedy}     & 96.5 & 87.1 & 86.2   \\ 
\texttt{Ours}     & 97.8 & 93.2 & 93.9  \\ \bottomrule
\end{tabular}
\caption{Recall rate of mentions of different nested depth on CoNLL-2012 dev set. There are 16873, 2100, and 182 mentions respectively of each nested depth.}
\label{tab:qualitative_depth}
\end{center}
\end{table}

\begin{table}[ht]
\begin{center}
\begin{tabular}{cccc}
\toprule
        & Avg. P        & Avg. R        & Avg. F1       \\ \midrule
\citeauthor{joshi-etal-2020-spanbert} (S)     & 81.6 & 78.6          & 80.1  \\
\texttt{Sigmoid} & 80.8          & 79.6          & 80.2          \\
\texttt{MTL}  & 80.8          & 78.9          & 79.8    \\
\texttt{Ours}    & 81.1          & 79.9 & \textbf{80.5} \\ \bottomrule
\end{tabular}
\caption{Comparison with three constructed baselines.}
\label{tab:result_test_conll_sigmoid}
\end{center}
\end{table}
\paragraph{\textbf{Widths of Mentions.}} 
We also compare the recall rate for mentions of different widths in \cref{tab:qualitative_width}. We show that our model can detect longer spans better, which are usually more difficult to detect. For spans with 5--12 words, our structured model still maintains a recall rate of 96.5\%, compared to a sharp drop for the greedy unstructured model.

\begin{table}
\begin{center}
\begin{tabular}{cccc}
\toprule
Span Width    &  1-4    &  5-12        &  12+      \\ \midrule
\texttt{Greedy}     & 96.2 &  92.7 &  82.5  \\ 
\texttt{Ours}     & 97.8 & 96.5  &  85.2 \\ \bottomrule
\end{tabular}
\caption{Recall rate of mentions of different width on the CoNLL-2012 dev set. There are 16356, 2180, and 619 mentions respectively of each width interval.}
\label{tab:qualitative_width} 
\end{center}
\end{table}

\subsection{Semantic Role Labeling}
For semantic role labeling, we report the precision, recall, and F1 score on the CoNLL-2012 SRL dataset. The gold predicates are provided during both training and evaluation. Therefore, the model has to focus on extracting the correct arguments and classifying their roles for each predicate. The results are shown in \cref{tab:result_test_srl} in comparison with previous span-based models. \citeauthor{he-etal-2018-jointly}\textsubscript{SpanBERT} refers to the model of \citet{he-etal-2018-jointly} with SpanBERT\textsubscript{large} \citep{joshi-etal-2020-spanbert} as a sentence encoder.

\begin{table}
\centering
\resizebox{\columnwidth}{!}{
\begin{tabular}{cccc}
\toprule
        & Avg. P        & Avg. R        & Avg. F1       \\ \midrule
\citet{he-etal-2018-jointly}     & - & - & 85.5  \\ 
\citet{ouchi-etal-2018-span}     & 87.1 & 85.3 & 86.2  \\ 
\citet{li2019dependency}     & 85.7 & 86.3 & 86.0  \\ 
\citet{DBLP:journals/corr/abs-1904-05255}     & 85.9 & 87.0 & 86.5  \\ \midrule
\citeauthor{he-etal-2018-jointly}\textsubscript{SpanBERT}     & 88.3 & 85.9 & 87.1  \\ 
\texttt{Ours}     & 88.1 & 86.9 & \textbf{87.5}  \\ \bottomrule
\end{tabular}
\caption{Results on the test set of the CoNLL-2012 semantic role labeling task. The precision, recall, and F1 scores are averaged over all semantic roles.
}
\label{tab:result_test_srl}
}
\end{table}

Next, we report the performance of our span selector on the SRL task. 
Following the same trend as coreference resolution, we find that our structured model is able to extract much more accurate arguments and thus, significantly reduce the memory consumption for the downstream task. 
While keeping a comparable recall rate of gold arguments (96.5\% for the greedy selector and 96.2\% for \texttt{ours}), our span selector reduces the number of enumerated spans by 21.2\%.
We compare the accuracy of retrieving unlabeled argument spans in \cref{tab:span_accuracy}. \texttt{BIO} refers to the tagger-style SRL model using the same text encoder. Our model outperforms both baselines.\looseness=-1

\begin{table}
\begin{center}
\begin{tabular}{cccc}
\toprule
            &  P    &  R & F1            \\ \midrule
\texttt{BIO}  &  90.2 & 91.0 &  90.6    \\ 
\texttt{Greedy}  &  92.8 & 90.2 &  91.5    \\ 
\texttt{Ours}  &  92.8 & 91.2 &  92.0   \\ \bottomrule
\end{tabular}
\caption{Comparison of unlabeled span accuracy.}
\label{tab:span_accuracy}
\end{center}
\end{table}

\subsection{Qualitative Examples}
In this section, we show a qualitative example to illustrate the grammar learned by our span selector for coreference resolution and SRL. 
Two sets of extracted spans for the same input sentence are shown in \cref{tab:qualitative_example}. For coreference resolution, our model selects maximal NPs (containing all modifiers) and verbs. While in SRL, the parse tree consists of much denser and syntactically heterogeneous spans of NPs, PPs, modal verbs, adverbs, etc. This comparison empirically shows that our model is capable and robust enough to learn a complex underlying grammar from partial annotation.

\subsection{Memory Efficiency}

\begin{table}%
\begin{center}
\begin{tabular}{ccc}
\toprule
            &  Coref    &  SRL            \\ \midrule
\texttt{Greedy}     & 11.5 &  12.2    \\ 
\texttt{Ours}     & 8.5 &  9.3   \\ \bottomrule
\end{tabular}
\caption{Comparison of peak GPU memory usage in GBs at inference time on the development set.}
\label{tab:mem_efficiency}
\end{center}
\end{table}
We further analyze the memory efficiency of our model. We evaluate the peak GPU memory usage on the development set of OntoNotes. For both tasks, we see a significant reduction in memory usage of 27\% for coreference and 24\% for SRL.

\begin{table}
\centering
\begin{tabular}{ll}
\toprule
Coref& \makecell[l]{[[The world' s] fifth [Disney] park]\\ will soon [open] to [the public] here .} \\ \midrule
SRL& \makecell[l]{[The world' s [fifth] [Disney] [park]]\\ \text{[will] [soon] [ [open] [to the public]]} \\ {[here] .}} \\ \bottomrule
\end{tabular}
\caption{A qualitative example of the grammar learned by the structured span selector.}
\label{tab:qualitative_example}
\end{table}
\section{Conclusion}
In this paper, we proposed a novel structured model for span selection. In contrast to prior span selection methods, the model is structured, which allows it to model spans better. Instead of a greedy span selection procedure, the span selector uses partial span annotations provided in data to directly obtain the set of optimal spans. 
We evaluated our span selector on two typical span prediction tasks, namely coreference resolution and semantic role labeling, and achieved consistent gains in terms of accuracy as well as efficiency over greedy span selection.

\section*{Ethical Considerations}
To the best of our knowledge, the datasets used in our work do not contain sensitive information, and we foresee no further ethical concerns with the work.\looseness=-1

\section*{Acknowledgements}
Mrinmaya Sachan acknowledges support from an ETH Z\"urich Research grant (ETH-19 21-1) and a grant from the Swiss National Science Foundation
(project \#201009) for this work.

\bibliography{anthology,custom}
\bibliographystyle{acl_natbib}

\clearpage
\appendix
\onecolumn

\section{A Weighted Context-Free Grammar}\label{app:wcfg}

\subsection{The Grammar}\label{app:grammar}
In our WCFG $\langle \Sigma, N, \Sstart, R, \rho \rangle$, $\Sigma$ is the set of all tokens in the vocabulary, $N = \{\Sstart, \Xsigma, \Xnotsigma \}$, where $\nt{S}$ is the start symbol, $\Xsigma$ is the span of interest, and $\Xnotsigma$ is the spans that are not of interest. The complete set of production rules $R$ is shown in \cref{tab:prod_rules}.

\begin{table}[h!]
\centering
    \begin{tabular}{ C C L } 
    \toprule
     \Sstart & \rightarrow & \Xsigma \,\Xsigma  \\
    \Sstart &\rightarrow & \Xsigma \,\Xnotsigma  \\
    \Sstart &\rightarrow & \Xnotsigma \,\Xsigma  \\
    \Sstart &\rightarrow & \Xnotsigma \,\Xnotsigma  \\ 
     \midrule
    \Xsigma &\rightarrow & \Xsigma \,\Xsigma  \\
    \Xsigma &\rightarrow & \Xsigma \,\Xnotsigma  \\
    \Xsigma &\rightarrow & \Xnotsigma \,\Xsigma  \\
    \Xsigma &\rightarrow & \Xnotsigma \,\Xnotsigma  \\
     \midrule
    \Xnotsigma &\rightarrow & \Xsigma \,\Xsigma  \\
    \Xnotsigma &\rightarrow & \Xsigma \,\Xnotsigma  \\
    \Xnotsigma &\rightarrow & \Xnotsigma \,\Xsigma  \\
    \Xnotsigma &\rightarrow & \Xnotsigma \,\Xnotsigma  \\
     \midrule
    \Xsigma & \rightarrow & x,\quad\quad \forall x \in \Sigma \\
    \Xnotsigma & \rightarrow & x,\quad\quad \forall x \in \Sigma \\
\bottomrule
    \end{tabular}
    \caption{Production rules $R$ of our WCFG.}
    \label{tab:prod_rules}
\end{table}
We only assign nontrivial weights to rules ${}_i\nt{X}_k \rightarrow {}_i \nt{Y}_j\,{}_j\nt{Z}_k$ where $X$ is $\Xsigma$. That is to say, in \cref{eq:parse_score}, $s_p({}_i\nt{X}_k \rightarrow {}_i \nt{Y}_j\,{}_j\nt{Z}_k) = 1$ where $X$ is not $\Xsigma$.

\subsection{The Inside--Outside Algorithm}
For a span $[i,\nt{X}, k]$, its inside value can be expressed as
\begin{align}\label{eq:io}
     \beta & ([i,\nt{X}, k]) = 
     \sum_{\nt{X} \rightarrow \nt{Y}\,\nt{Z} \, \in R} \Bigg( \exp s_p([i, \nt{X}, k]) \times 
    \Big( \sum_{j=i+1}^{k-1}\beta([i,\nt{Y}, j]) \times \beta([j, \nt{Z}, k]) \Big) \Bigg)
\end{align} %
In the case when $k=i+1$, we have $\beta([i,\nt{X}, k]) = \exp \scorep{[i, \nt{X}, k]} $. %
The inside value of the entire sentence $\beta([0, \nt{S}, |{\bf w}|])$ %
is exactly $Z$, the sum of scores of all parses.

\subsection{The CKY Algorithm}\label{app:cky}
We use the CKY algorithm to find an optimal parse tree $t^* \in \mathcal{T}(\bw)$ that maximizes $\rho(t)$. The recursive function when $i < k$ used is: 
\begin{align}\label{eq:cky}
        \gamma & ([i,\nt{X},k]) = 
        \max_{\nt{X} \rightarrow \nt{Y}\,\nt{Z} \, \in R} \Bigg\{ s_p([i, \nt{X}, k])  +
         \max_{i < j < k} \Big\{ \gamma([i,\nt{Y},j]) + \gamma([j,\nt{Z},k]) \Big\} \Bigg\}
\end{align} %
In the case when $k=i+1$, we have $\gamma ([i,\nt{X},k]) = s_p([i,\nt{X},k])$. 

\subsection{A Derivation of \cref{eq:anchored_prob}}\label{app:ioisfb}
\begin{subequations}
\begin{align}
   \frac{\partial \log Z}{\partial s_p(\sigma_{ik})} &= \frac{1}{Z} \times \frac{\partial Z}{\partial s_p(\sigma_{ik})} \\
    &= \frac{1}{Z} \times \sum_{t \in \mathcal{T}(\bw)} \frac{\partial \rho(t) }{\partial s_p(\sigma_{ik})}\\
    &= \frac{1}{Z} \times \sum_{t \in \mathcal{T}_{ik}(\bw)} \frac{\partial \rho(t) }{\partial s_p(\sigma_{ik})}\\
    &= \frac{1}{Z} \times \sum_{t \in \mathcal{T}_{ik}(\bw)} \left( \frac{\partial \exp s_p(\sigma_{ik})}{\partial s_p(\sigma_{ik})} \times \prod_{\sigma \in t \backslash \{\sigma_{ik}\}}  \exp s_p(\sigma) \right)\\
    &= \frac{1}{Z} \times \sum_{t \in \mathcal{T}_{ik}(\bw)} \left( \exp s_p(\sigma_{ik})  \times \prod_{\sigma \in t \backslash \{\sigma_{ik}\}}  \exp s_p(\sigma) \right)\\
    &= \frac{1}{Z} \times \sum_{t \in \mathcal{T}_{ik}(\bw)} \rho(t) \\
    &= p(\sigma_{ik} \mid \bw)
\end{align}
\end{subequations}

\section{Experimental Settings}
The systems are implemented with PyTorch. We use SpanBERT$_{\text{large}}$ as text encoder. We train the model for 20 epochs and select the best-performing model on the development set for testing. The documents are split into 512 word segments to fit in SpanBERT$_{\text{large}}$. Models used for coreference resolution have 402 million learnable parameters, and models for SRL have 382 million learnable parameters. 
We closely follow the hyperparameter settings of \citet{joshi-etal-2020-spanbert} and build our models upon the codebase of \citet{xu-choi-2020-revealing}\footnote{\url{https://github.com/lxucs/coref-hoi}} under Apache License 2.0. The learning rate of SpanBERT$_{\text{large}}$ parameters is set to $1 \times 10^{-5}$ with $0.01$ decay rate, and the learning rate of task parameters is set to $3 \times 10^{-4}$. The dropout rate of feedforward neural network scorers is set to $0.3$.
When training our model, we randomly sample $0.1|D|$ negative spans that are not mentions and add their negative log-likelihood $-\log p(\sigma_{ij})$ to the training objective. This is to prevent $p(\sigma_{ij})$ from converging to $1$. 
For SRL task, we use a batch size of 32 for 40 epochs and the same learning rate with coreference resolution. The same negative sampling technique is applied.
Our models are trained on Nvidia Tesla V100 GPUs with 32GB memory. The average training time is around 8 hours for \citeauthor{joshi-etal-2020-spanbert} (S) baseline and around 9 hours for our model. For SRL models, training takes 25 hours. 

\section{Results on the Development Set}
In this section, we report the results that our models get on the development sets of OntoNotes and LitBank.

\begin{table}
\begin{center}
\begin{tabular}{cccc}
\toprule
        & Avg. P        & Avg. R        & Avg. F1       \\ \midrule
\citeauthor{joshi-etal-2020-spanbert} (S)     & 82.0 & 78.8          & 80.4  \\
\texttt{Sigmoid} & 81.6          & 79.4          & 80.5          \\
\texttt{Ours}    & 81.1          & 80.2 & \textbf{80.7} \\ \bottomrule
\end{tabular}
\caption{Results on CoNLL-2012 coreference resolution development set.}
\label{tab:result_dev_conll}
\end{center}
\end{table}

\begin{table}
\begin{center}
\begin{tabular}{cccc}
\toprule
        & Avg. P        & Avg. R        & Avg. F1       \\ \midrule
\citeauthor{joshi-etal-2020-spanbert} (S)     & 78.8 & 73.7          & 76.1  \\
\texttt{Ours}    & 77.4          & 75.6 & \textbf{76.6} \\ \bottomrule
\end{tabular}
\caption{Results on LitBank development set.}
\label{tab:result_dev_litbank}
\end{center}
\end{table}

\begin{table}[ht!]
\begin{center}
\begin{tabular}{cccc}
\toprule
        & Avg. P        & Avg. R        & Avg. F1       \\ \midrule
\citeauthor{he-etal-2018-jointly}\textsubscript{SpanBERT}     & 87.9 & 85.0 & 86.4  \\ 
\texttt{Ours}    & 87.8          & 86.2 & \textbf{87.0} \\ \bottomrule
\end{tabular}
\caption{Results on CoNLL-2012 semantic role labeling development set.}
\label{tab:result_dev_srl}
\end{center}
\end{table}

\end{document}